\documentclass[final]{opt2026}
\usepackage{booktabs, array,graphicx,microtype,enumitem,mathtools, multirow,xcolor,tikz, float}
\usepackage{comment}
\usetikzlibrary{arrows.meta, positioning, fit, backgrounds}
\title[PDE-State Conditioning]{PDE-State Conditioning for Learned Optimization Using \\ Physics-Informed Neural Networks}
\optauthor{
\Name{Farzana Yasmin} \Email{farzanayasmin.cs@gmail.com}\\
\addr Department of Computer Science, University of Houston, Houston, TX, USA 
\AND
\Name{Ricardo Vilalta} \Email{rvilalta@uaustin.org}\\
\addr Center for Science, University of Austin, Austin, TX, USA}

\newcommand{\LPDE}{\mathcal{L}_{\mathrm{PDE}}}
\newcommand{\LIC}{\mathcal{L}_{\mathrm{IC}}}
\newcommand{\LBC}{\mathcal{L}_{\mathrm{BC}}}
\newcommand{\Method}{\textsc{PDE-State(G2)+FiLM}}
\newcommand{\BaseLO}{\textsc{Unconditioned LO}}
\newcommand{\eps}{\varepsilon}

\begin{document}
\maketitle
\begin{abstract}
Physics-informed neural network (PINN) optimization depends on global properties of the underlying PDE and how the physics residual evolves across the domain. Standard optimizers such as Adam use fixed gradient-based statistics, while learned PINN optimizers replace the fixed rule with a learned update built from local parameters and optimization-history features. However, none explicitly use PDE-level information. We study two coupled design questions: \emph{what PDE-level information should a learned optimizer observe, and how should that information influence its update?} We introduce an 11-dimensional PDE state capturing operator structure, residual geometry, and short-horizon update response, and use it to condition a structured meta-learned optimizer through bounded feature-wise modulation.  Repeated design studies identify the compact residual-geometry state G2 and bounded FiLM as the strongest overall choices among those tested. The resulting PDE-State-conditioned optimizer improves all evaluation metrics (relative-$L_2$, $L_1$, solution MSE, and physics MSE) over Adam and the unconditioned learned optimizer on Linear Advection, KdV, and Burgers. 
\end{abstract}
\begin{keywords}learned optimization, physics-informed neural networks, meta-learning, PDE-state conditioning, optimizer robustness\end{keywords}

\section{Introduction}\label{sec:introduction}
Physics-informed neural networks (PINNs) approximate solutions to Partial Differential Equations (PDEs) by training a neural network to satisfy the underlying physics of the governing equation along with its initial and boundary conditions \citep{raissi2019pinns,karniadakis2021physics}. This makes the training quite different from ordinary supervised learning, although some optimization problems still exist \citep{wang2021gradflow,wang2022ntk,krishnapriyan2021failure,wuw2023sampling}. One avenue to improve on PINNs is to use learned optimizers (LOs) to replace hand-designed update rules with learned ones across related training tasks \citep{andrychowicz2016learning,wichrowska2017scale,harrison2022closer}; this idea has been applied through meta-learned optimization \citep{bihlo2024pinn}. We hypothesize that this combination of PINNs and LOs—a cross-PDE transfer—improves when the learned update can see the current PDE training state, while the baseline optimizer structure and its local inputs remain intact. 

Our work raises two design questions on the interaction between PINNs and LOs: (i) \emph{which} PDE-level signals are informative beyond local optimizer history, and (ii) \emph{how} should they act on the learned update? We address both questions experimentally. We construct an 11-dimensional PDE state that combines coarse operator descriptors, residual geometry, and feedback from the previous learned correction. A small context network maps these states to affine scale and shift parameters \citep{perez2018film}, which then modulate the optimizer's hidden representation. The final model carries its gains beyond the Linear Advection meta-training problem to KdV and Burgers without re-meta-training. Its aggregate performance beats Adam and an unconditioned learned optimizer across all reported metrics, with lower variability across independently learned optimizer checkpoints.


\section{PINN Formulation}
\label{sec:background}

Consider a time-dependent PDE on $\Omega\subset\mathbb{R}^d$ and $t\in[0,T]$,
\begin{subequations}
\label{eq:pde_system}
\begin{align}
\mathcal{N}_{\lambda}[u](t,x)&=0,
&& (t,x)\in(0,T]\times\Omega,
\label{eq:pde_governing}\\
\mathcal{B}_{\lambda}[u](t,x)&=g(t,x),
&& (t,x)\in(0,T]\times\partial\Omega,
\label{eq:pde_boundary}\\
u(0,x)&=u_{0,\lambda}(x),
&& x\in\Omega.
\label{eq:pde_initial}
\end{align}
\end{subequations}
Here, $\mathcal{N}_\lambda$ denotes the governing PDE operator, $\mathcal{B}_\lambda$ the boundary operator, and $\lambda$ collects PDE parameters such as advection speed, viscosity, or dispersion. A PINN $u_\theta(t,x)$ parameterizes the unknown solution $u$, where $\theta$ denotes the trainable network parameters. Automatic differentiation evaluates the derivatives in $\mathcal{N}_\lambda$ and $\mathcal{B}_\lambda$, and $\theta$ is learned by minimizing the residuals of the governing PDE along with the initial and boundary conditions.

\noindent
For interior collocation points $\mathcal{D}_f=\{(t_f^{(j)},x_f^{(j)})\}_{j=1}^{N_f}$, the PDE residual and loss are 
$r_{\rm PDE}(t,x;\theta,\lambda) = \mathcal{N}_\lambda[u_\theta](t,x)$, and 
$\LPDE(\theta) = \frac1{N_f}\sum_{j=1}^{N_f}\left|r_{\rm PDE}(t_f^{(j)},x_f^{(j)};\theta,\lambda)\right|^2$.
For initial-condition samples,
$r_{\rm IC}(x;\theta,\lambda) = u_\theta(0,x)-u_{0,\lambda}(x)$, and  
$\LIC(\theta) = \frac1{N_i}\sum_{j=1}^{N_i}|r_{\rm IC}(x_i^{(j)};\theta,\lambda)|^2$.
For boundary samples,
$r_{\rm BC}(t,x;\theta,\lambda) = \mathcal{B}_\lambda[u_\theta](t,x)-g(t,x)$, and 
$\LBC(\theta) = \frac1{N_b}\sum_{j=1}^{N_b}|r_{\rm BC}(t_b^{(j)},x_b^{(j)};\theta,\lambda)|^2$.
The usual soft-constrained PINN objective is
\begin{equation}
\mathcal{L}(\theta)=\gamma_f\LPDE(\theta)+\gamma_i\LIC(\theta)+\gamma_b\LBC(\theta).
\label{eq:pinn_obj}
\end{equation}

The PINN parameters are therefore obtained from $\theta^\star=\arg\min_{\theta}\mathcal{L}(\theta)$. For our one-dimensional benchmarks on $x\in[-1,1]$, we encode the spatial coordinate as
$\phi(x)=[\cos(\pi x),\sin(\pi x)]$. Since $\phi(x+2)=\phi(x)$, the PINN representation is periodic in $x$, so we enforce periodicity by construction rather than through a boundary-loss term. The active soft constraints are therefore primarily $\mathcal{L}_{\mathrm{PDE}}$ and $\mathcal{L}_{\mathrm{IC}}$.


\section{PDE-state-conditioned learned optimization}
\label{sec:method}

\subsection{Structured learned-optimizer baseline}
\label{sec:learned_optimizer_baseline}

Following the structured meta-learned optimizer form used by Bihlo \citep{bihlo2024pinn}, we apply the learned update to each PINN parameter, 
\begin{equation}
\theta_{i,t+1}=\theta_{i,t}-f(z_{i,t};\vartheta),
\label{eq:learned_parameter_update}
\end{equation}
Here, $\theta_{i,t}$ is the $i$-th scalar PINN parameter at optimization step $t$, $z_{i,t}$ is its local optimizer-feature vector, and $f(\cdot;\vartheta)$ is the learned update function parameterized by the meta-learned optimizer weights $\vartheta$. Thus, $\theta$ denotes the PINN parameters being optimized, whereas $\vartheta$ denotes the optimizer's learned parameters. The structured update function is
\begin{equation}
f(z_{i,t};\vartheta)
=\exp\!\left(\lambda_3 s^{\rm nom}_{i,t}\right)w^{\rm Adam}_{i,t}
+\frac{\lambda_1 d^{\rm BB}_{i,t}\exp\!\left(\lambda_2 s^{\rm BB}_{i,t}\right)}{\sqrt{v_{i,t}}+\eps}.
\label{eq:structured_learned_update}
\end{equation}
The optimizer network $f(\cdot;\vartheta)$ produces three output heads, $s^{\rm nom}_{i,t}$, $d^{\rm BB}_{i,t}$, and $s^{\rm BB}_{i,t}$. Here, $w^{\rm Adam}_{i,t}
=\eta\widehat m_{i,t}/(\sqrt{\widehat v_{i,t}}+\eps)$ is the nominal Adam step.
The head $s^{\rm nom}_{i,t}$ rescales the nominal Adam contribution, while $d^{\rm BB}_{i,t}$ and $s^{\rm BB}_{i,t}$ determine the direction and magnitude, respectively, of the learned correction. The input $z_{i,t}$ contains local parameter, gradient, moment, scale, and time features. The update rule in Equations~\eqref{eq:learned_parameter_update}--\eqref{eq:structured_learned_update} remains fixed across all conditioned variants and \BaseLO{}. Only how PDE-state information enters the meta-learned optimizer changes. Further implementation details are in Appendix~\ref{app:optimpl}.

\subsection{PDE state: global information complementary to local history}
\label{sec:pde_state}

For each PINN parameter $\theta_i$, the meta-learned optimizer receives a local input $z_{i,t}$. These are used to construct the PINN parameter update in Eq.~\eqref{eq:learned_parameter_update} while optimizing the PINN objective in Eq. ~\eqref{eq:pinn_obj}. Because local optimization history alone may not capture different training situations, we introduce a shared PDE state $s_t\in\mathbb{R}^{11}$ that summarizes both the governing-equation structure and the current physics-training state. The full 11-dimensional vector is a candidate information space, divided into three groups by role. G1 provides information on coarse operator structure; G2 describes how the current PDE residual is distributed in space--time beyond its mean squared value; and G3 shows what the previously learned black-box displacement does to the PDE and IC residuals. Table~\ref{tab:state_roles} summarizes the feature roles, and Appendix~\ref{app:state} gives exact definitions.

\begin{table}[!h]
\centering\small
\caption{PDE-state information supplied to the learned optimizer. Exact formulas and probe details are given in Appendix~\ref{app:state}.}
\label{tab:state_roles}
\begin{tabular}{@{}clp{0.63\linewidth}@{}}
\toprule
Group & Dim. & Role in optimizer state \\
\midrule
G1 & 7 & Operator structure: time dependence, derivative orders, nonlinearity, advection, diffusion, dispersion.\\
G2 & 2 & Residual geometry: concentration of squared PDE residual energy and local residual roughness.\\
G3 & 2 & Update response: PDE and IC residual response to the previous learned black-box displacement.\\
\bottomrule
\end{tabular}
\end{table}


\subsection{PDE-state conditioning of the optimizer representation} 
\label{sec:bounded_conditioning}

\begin{figure}[t]
\centering
\includegraphics[width=0.98\linewidth]{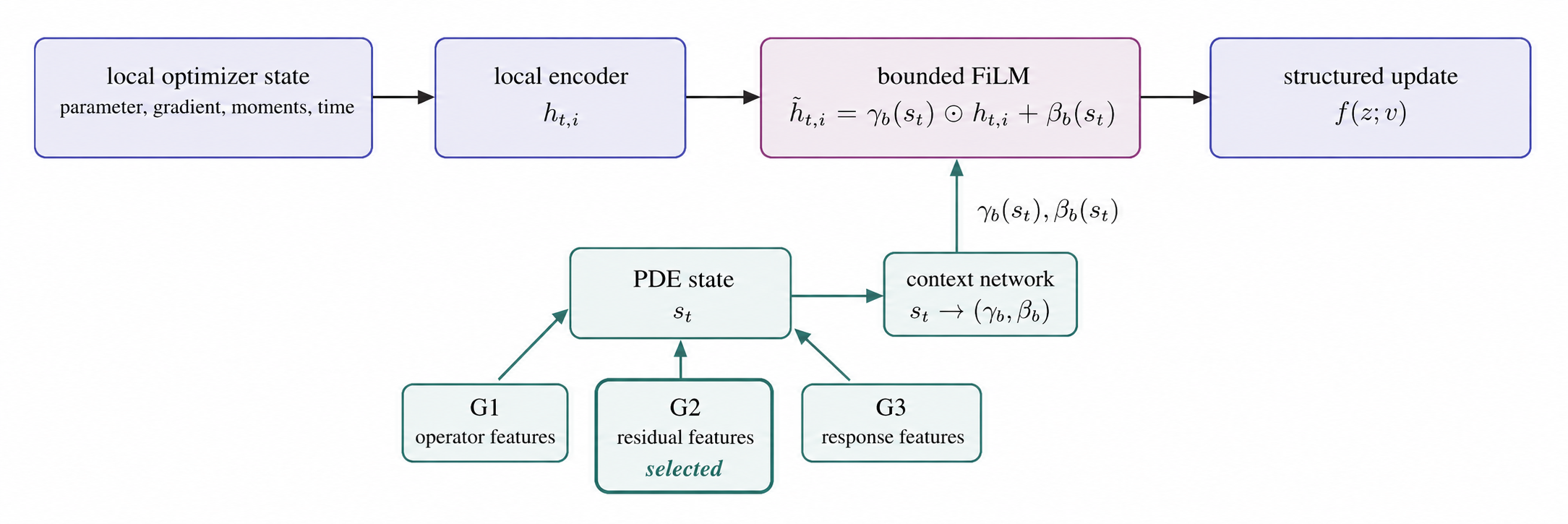}
\caption{PDE-state-conditioned meta-learned optimizer. The candidate PDE-state $s_t$ is formed from feature groups--operator features (G1), residual features (G2), and update-response features (G3). The state study identifies G2 as the strongest among the tested groups. A shared context network maps the selected PDE state to the bounded FiLM parameters $\gamma_b(s_t)$ and $\beta_b(s_t)$, which modulate the local hidden representation.}
\label{fig:method}
\end{figure}

To learn the PINN optimization update rule, we treat the local optimizer input and the global PDE state separately. The meta-learned optimizer first maps the local input $z_{i,t}$ to a hidden representation $h_{i,t}=F_\phi(z_{i,t})$, while a small context network maps $s_t$ to feature-wise scale and shift parameters that condition this hidden representation. Figure~\ref{fig:method} shows the full architecture. Standard FiLM \citep{perez2018film} applies an affine transformation as follows:
\begin{equation}
\operatorname{FiLM}(h\mid s)=\gamma(s)\odot h+\beta(s).
\label{eq:film}
\end{equation}
In our optimizer, we apply this conditioning to the output of the first hidden layer, before the second hidden layer and the update heads. We use a bounded, identity-centered form of standard FiLM,
\begin{align}
\gamma_b(s_t)&=1+\rho\tanh(\gamma_{\rm raw}(s_t)),\\
\beta_b(s_t)&=\rho\tanh(\beta_{\rm raw}(s_t)),\\
\widetilde h_{i,t}&=\gamma_b(s_t)\odot h_{i,t}+\beta_b(s_t),\qquad \rho=0.10.
\label{eq:bounded_film}
\end{align}
Hence $\gamma_b\in[0.9,1.1]$ and $\beta_b\in[-0.1,0.1]$. Because the scale is centered at one, the PDE state modifies the original hidden representation rather than replacing it. Bounding the scale and shift limits how much the global controller can perturb the learned optimizer. We evaluate whether this approach to introducing PDE-state information is necessary through several ablation studies. Appendix~\ref{app:conditioning} summarizes the conditioning alternatives considered, including direct input, restricted scale/shift variants, structured and centered modulation, and higher-capacity alternatives. Quantitative comparisons are reported in Section~\ref{sec:conditioning_results} and Appendix \ref{app:conditioning}.

\section{Results and analysis}\label{sec:results}
\subsection{Accuracy, physics satisfaction and robustness across source and transfer PDEs}

All learned optimizers are meta-trained only on Linear Advection and evaluated on LA, KdV, and Burgers without re-meta-training. The principal comparison uses three independent meta-training seeds, with five meta-test runs per checkpoint and 1,000 PINN training epochs per evaluation. Full experimental details are given in Appendix~\ref{app:optimpl}.
Table~\ref{tab:headline_allmetrics} summarizes the main comparison among \Method, a standard Adam optimizer, and \BaseLO. Across LA, KdV, and Burgers, \Method{} has the lowest aggregate mean among the compared optimizers for all four evaluation metrics. On LA, conditioning produces the largest gain over Adam and \BaseLO{}, while Burgers gives the narrowest margin over Adam. The mean relative-$L_2$ reductions are 74.3\%, 30.0\%, and 7.3\% versus Adam and 68.5\%, 25.9\%, 24.1\% versus \BaseLO{} on LA, KdV, and Burgers, respectively. The same overall advantage is also observed for $L_1$, solution MSE, and physics MSE on all three PDEs.

\begin{table}[t]
\centering\scriptsize
\caption{Performance after 1,000 PINN epochs. Five meta test runs are averaged within each block; entries are the mean $\pm$ SD across three blocks. For the learned optimizers, each block
corresponds to an independently meta-trained checkpoint. Lower is better; the best mean for each PDE/metric is bold.}
\label{tab:headline_allmetrics}
\begin{tabular}{@{}llcccc@{}}
\toprule
Method & PDE & Rel. $L_2$ & $L_1$ & Solution MSE & Physics MSE\\
\midrule
Adam & LA & .5200$\pm$.0224 & .3007$\pm$.0130 & .14088$\pm$.00840 & .04156$\pm$.00182\\
& KdV & .4648$\pm$.0147 & .2211$\pm$.0057 & .10952$\pm$.00685 & .03523$\pm$.00169\\
& Burgers & .1286$\pm$.0017 & .0425$\pm$.0019 & .00657$\pm$.00023 & .01102$\pm$.00103\\
\midrule
\BaseLO & LA & .4249$\pm$.1808 & .2468$\pm$.1058 & .11532$\pm$.07294 & .03451$\pm$.01669\\
& KdV & .4392$\pm$.1151 & .2110$\pm$.0475 & .10337$\pm$.05361 & .03378$\pm$.01347\\
& Burgers & .1570$\pm$.1280 & .0582$\pm$.0493 & .01451$\pm$.02044 & .02311$\pm$.02727\\
\midrule
\textbf{\Method{}} & LA & $\mathbf{.1339\pm.0855}$ & $\mathbf{.0770\pm.0497}$ & $\mathbf{.01398\pm.01555}$ & $\mathbf{.00908\pm.00584}$\\
& KdV & $\mathbf{.3255\pm.0439}$ & $\mathbf{.1633\pm.0172}$ & $\mathbf{.05457\pm.01557}$ & $\mathbf{.02107\pm.00164}$\\
& Burgers & $\mathbf{.1192\pm.0278}$ & $\mathbf{.0304\pm.0027}$ & $\mathbf{.00643\pm.00301}$ & $\mathbf{.00655\pm.00130}$\\
\bottomrule
\end{tabular}
\end{table}


Repeated meta-tests on a single checkpoint capture test-time variability, but not sensitivity to meta-training initialization. After averaging meta-test runs within each meta-training, the standard deviation (SD) across independently meta-trained \Method{} checkpoints is lower than that of \BaseLO{} on all twelve PDE--metric combinations. For relative $L_2$, the SD is reduced by 52.7\% (LA), 61.9\% (KdV), and 78.3\% (Burgers). Across all four metrics, the reduction ranges from 52.7\% to 95.2\%. We therefore interpret robustness here as reduced sensitivity to learned-optimizer meta-training initialization relative to the unconditioned learned baseline.

\subsection{What PDE-state information is useful?}
\label{sec:state_results}
To explore which information is useful as the global state, we keep the conditioning fixed and first compare the performance of G1, G2, G3, their pairwise combinations, and All11. We then repeat four representative configurations--G2, G1+G3, G2+G3, and All11-- for each PDE across different meta-training runs. Across the 12 PDE--metric combinations reported in Table~\ref{tab:group_allmetrics_agg} in Appendix~\ref{app:feature_screen}, G2 has the lowest aggregate mean in 10. The exceptions are KdV physics MSE, where G1+G3 is slightly lower, and Burgers relative-$L_2$, where All11 is slightly lower. In a direct comparison with All11, G2 is lower in 11 of 12 combinations. Figure~\ref{fig:design_studies}(a) shows the corresponding relative-$L_2$ results of the four group comparisons.
The repeated strength of G2 motivated several residual-focused follow-ups. G2-ext, a variant with additional residual-focused features, is worse than G2 on all three PDEs. G2-balance, with gradient balance features, improves the four KdV metrics in its tested run but does not improve LA or Burgers. ResGeomBasis also performs poorly. These results indicate that simply adding more residual or PDE-state information is not automatically beneficial. Among the representations tested, the compact concentration and roughness state G2 provides the strongest overall result. We therefore retain G2 as the fixed PDE-state representation for the final comparison with Adam and \BaseLO{}. Full feature study ablations are in Appendix~\ref{app:feature_groups}.

\subsection{How the PDE state enters the optimizer matters}
\label{sec:conditioning_results}
\begin{figure}[t]
\centering
\includegraphics[width=0.80\linewidth]{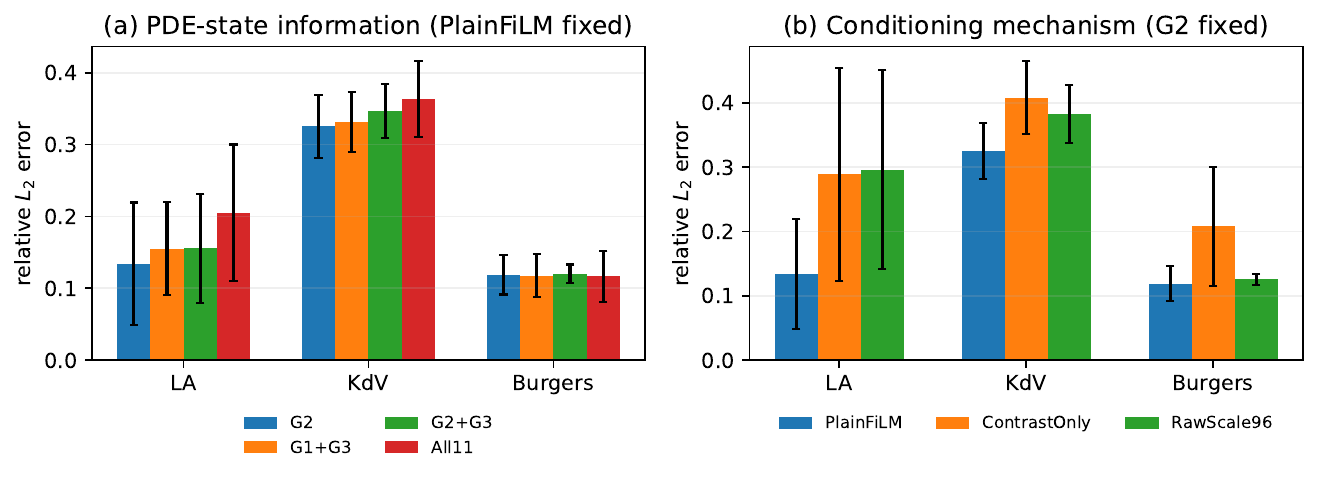}
\caption{Relative-$L_2$ error performance comparison. (a) PDE-state information with conditioning (PlainFiLM) fixed. (b) Conditioning with features (G2) fixed in an independent re-meta-training study.}
\label{fig:design_studies}
\end{figure}

 We next examine whether the way PDE-state information enters the optimizer changes the result. We designed most initial conditioning controls with All11 fixed, including direct-input, restricted FiLM, structured/centered, and capacity controls. Several alternatives look excellent on a single run, for example, ContrastOnly on LA/Burgers, Groupwise FiLM, and RawScale96 at one meta-run. But their advantage was not consistent after independent re-meta-training. Once G2 was selected as the strong-performing group from the feature study, we therefore repeated the key conditioning comparisons, such as plain FiLM, centering, and capacity control, with G2 fixed.  
 Table~\ref{tab:conditioning_comparison} shows that the preferred conditioning mechanism also depends on the state representation. With All11 fixed, RawScale96 has the lowest mean relative-$L_2$ on KdV, while PlainFiLM is lowest on LA and Burgers. With G2 fixed, PlainFiLM is lowest on all three PDEs. The full four-metric comparison (Table~\ref{tab:conditioning_g2_full} and Table~\ref{tab:conditioning_all11} in Appendix~\ref{app:conditioning}) strengthens this result. With G2 fixed, PlainFiLM has the lowest aggregate mean among the three mechanisms in all 12 PDE--metric comparisons. Figure~\ref{fig:design_studies}(b) shows the corresponding relative-$L_2$ results with G2 fixed. Appendix~\ref{app:conditioning} contains all details of these experiments and their corresponding results.

\begin{table}[t]
\centering
\small
\caption{Repeated conditioning comparison with All11 or G2 held fixed. Entries are relative-$L_2$ mean $\pm$ SD across three independently meta-trained checkpoints after averaging five meta test runs within each checkpoint. Lower is better; boldface marks the lowest mean within each state/PDE.}
\label{tab:conditioning_comparison}
\begin{tabular}{@{}llccc@{}}
\toprule
PDE state & Conditioning & LA & KdV & Burgers \\
\midrule
All11 & PlainFiLM & $\mathbf{.1365\pm.0896}$ & .3445$\pm$.0320 & $\mathbf{.1407\pm.0395}$ \\
& ContrastOnly & .2933$\pm$.1569 & .3957$\pm$.0565 & .2120$\pm$.1485 \\
& RawScale96 & .2281$\pm$.2294 & $\mathbf{.3289\pm.1177}$ & .1746$\pm$.1134 \\
\midrule
G2 & PlainFiLM & $\mathbf{.1339\pm.0855}$ & $\mathbf{.3255\pm.0439}$ & $\mathbf{.1192\pm.0278}$ \\
& ContrastOnly & .2890$\pm$.1659 & .4083$\pm$.0561 & .2079$\pm$.0930 \\
& RawScale96 & .2962$\pm$.1548 & .3829$\pm$.0451 & .1256$\pm$.0087 \\
\bottomrule
\end{tabular}
\end{table}

\section{Conclusion}
\label{sec:conclusion}

In this work, we studied how to incorporate PDE-level information into a meta-learned optimizer for PINNs, focusing on what information to expose and how it should condition the learned update. Across Linear Advection, KdV, and Burgers, PDE-state conditioning improves the final optimizer over both Adam and the \BaseLO{} on all four reported error metrics. It also reduces sensitivity to meta-training initialization relative to the unconditioned learned baseline. 
Taken together, the results suggest that learned optimization for PINNs depends on both the PDE-level information available to the optimizer and the way that information enters the parameter-level update process. The present study is limited to meta-training on Linear Advection and transfer to KdV and Burgers. Evaluating the same design principles across broader PDE families, alternative PDE-level features, and other conditioning mechanisms is an important direction for future research.

\clearpage
\appendix

\section{PDE-state representation}
\label{app:state}
The candidate 11-feature composed PDE state follows the fixed coordinate order summarized below. The final model activates only G2; we retain the remaining coordinates here because they define the feature ablations.
\begin{table}[H]
\centering\small
\caption{Coordinate layout of the 11-dimensional PDE state.}
\label{tab:pde_state_coordinates}
\begin{tabular}{@{}clp{0.67\linewidth}@{}}
\toprule
Indices & Group & Coordinates \\
\midrule
0--6 & G1 & time dependence; maximum temporal derivative order; maximum spatial derivative order; nonlinearity; advection; diffusion; dispersion \\
7--8 & G2 & $C_{\rm PDE}$ (residual concentration), $R_{\rm PDE}$ (residual roughness) \\
9--10 & G3 & $a^{\rm BB}_{\rm PDE}$, $a^{\rm BB}_{\rm IC}$ (response to the previous black-box correction) \\
\bottomrule
\end{tabular}
\end{table}

\subsection{G1 operator descriptors}
\label{app:g1}
\begin{table}[H]
\centering\small
\caption{Static operator descriptors used in PDEState11.}
\label{tab:g1_descriptors}
\begin{tabular}{@{}lccccccc@{}}
\toprule
Equation & Time & $d_t$ & $d_x$ & Nonlinear & Adv. & Diff. & Disp.\\
\midrule
Linear Advection & 1&1&1&0&1&0&0\\
Burgers & 1&1&2&1&1&1&0\\
KdV & 1&1&3&1&1&0&1\\
\bottomrule
\end{tabular}
\end{table}
Because meta-training uses LA only, these coordinates provide coarse structural information that distinguishes the transfer equations from the source PDE, but they don't establish that the optimizer has learned the meaning of diffusion or dispersion across PDEs.

\subsection{G2 residual concentration and roughness}
\label{app:g2}
For PDE-probe residuals $r_1,\ldots,r_N$, concentration is
\begin{equation}
C_{\rm PDE}=\frac{\sum_{i\in I_k}r_i^2}{\sum_{i=1}^N r_i^2+\eps},\qquad k=\max(1,\lceil0.20N\rceil),
\label{eq:residual_concentration}
\end{equation}
where $I_k$ contains the $k$ largest squared residuals. A larger value means a few probe points have concentrated residual energy.

Residual roughness compares each probe point to its nearest neighbor. If $z_i=(t_i,x_i)$ and $j(i)$ is the nearest neighboring probe point,
\begin{equation}
R_{\rm PDE}=\log\!\left[1+\frac1N\sum_{i=1}^N\frac{|r_i-r_{j(i)}|}{\|z_i-z_{j(i)}\|_2+\eps}\right].
\label{eq:residual_roughness}
\end{equation}
It is therefore large when nearby residual values change rapidly in space/time and small for a smoother residual field.

\subsection{G3 response to the previous black-box correction}
\label{app:g3}
Since the optimizer applies updates as $\theta_{t+1} = \theta_t-u_t$, the actual parameter displacement produced by the BB-only learned correction is $\Delta\theta^{\rm BB}=-u^{\rm BB}$. The probe recomputes the residual vectors under that displacement,
\begin{align}
\delta r^{\rm BB}_{\rm PDE}&=r_{\rm PDE}(\theta+\Delta\theta^{\rm BB})-r_{\rm PDE}(\theta), \label{eq:g3_delta_pde}\\
\delta r^{\rm BB}_{\rm IC}&=r_{\rm IC}(\theta+\Delta\theta^{\rm BB})-r_{\rm IC}(\theta), \label{eq:g3_delta_ic}
\end{align}
and stores
\begin{align}
a^{\rm BB}_{\rm PDE}&=-\cos(r_{\rm PDE},\delta r^{\rm BB}_{\rm PDE}),\label{eq:g3_alignment_pde}\\
a^{\rm BB}_{\rm IC}&=-\cos(r_{\rm IC},\delta r^{\rm BB}_{\rm IC}).\label{eq:g3_alignment_ic}
\end{align}
A positive value indicates that the residual change is anti-aligned with the current residual and therefore tends to reduce it. A negative value indicates movement in the opposite direction. If either norm in a cosine is effectively zero, the implementation returns zero.

The dynamic G2 and G3 coordinates are computed from residuals on fixed PDE and IC probe sets. Probe construction, refresh, smoothing, and normalization are described in Appendix~\ref{app:probe_refresh}

\section{Experimental setup and implementation}
\label{app:optimpl}
\label{sec:experiments}

\subsection{Source and transfer PDEs}
\label{app:pdes}
All learned optimizers are meta-trained only on one-dimensional Linear Advection (LA) and then evaluated on LA, KdV, and Burgers without re-meta-training. The three equations deliberately change the physical and differential structure seen by the optimizer:
\begin{align}
\text{LA:}\quad &u_t+c u_x=0, && c\sim U(0.9,1.1)\ \text{during meta-training},\\
\text{KdV:}\quad &u_t+u u_x-0.0025\,u_{xxx}=0,\\
\text{Burgers:}\quad &u_t+u u_x-\frac{0.01}{\pi}u_{xx}=0.
\label{eq:benchmark_pdes}
\end{align}
LA models advection of a quantity $u(x,t)$ at a constant velocity $c$. It provides a comparatively simple source family by varying the advection speed. KdV introduces nonlinearity and a third-order dispersive term, while Burgers combines nonlinear transport with weak second-order diffusion and can develop sharp solution gradients. All problems use $x\in[-1,1]$ with the periodic input embedding $[\cos(\pi x),\sin(\pi x)]$. LA uses $t\in[0,3]$; KdV and Burgers use $t\in[0,1]$. The initial conditions are $u(0,x)=-\sin(\pi x)$ for LA, $u(0,x)=\cos(\pi x)$ for KdV, and $u(0,x)=-\sin(\pi x)$ for Burgers.

\subsection{PINN and learned-optimizer configuration}
\label{app:model_config}

The PINN contains six 20-unit tanh hidden layers. Each task contains 10,000 PDE collocation points and 100 initial-condition points. The learned optimizer uses two 32-unit swish hidden layers.

Equation~\eqref{eq:structured_learned_update} is the structured meta-learned optimizer update used throughout the study. The implementation uses $\eta=\lambda_1=2\times10^{-4}$ and $\lambda_2=\lambda_3=10^{-3}$.
The local numeric feature vector contains the current parameter and gradient, four first moments, four second moments, the reciprocal square roots of the four second moments, and four moment-ratio features. The first/second-moment decay rates are $\{0.5,0.9,0.99,0.999\}$. Numeric features are normalized by root mean square before appending the time embedding
\begin{equation}
\tanh(t/x),\qquad x\in\{1,3,10,30,100,300,1000,3000,10000,30000,100000\}.
\label{eq:time_embedding}
\end{equation}
In the conditioned model, the PDE state acts after the first hidden layer. The context network has a 16-unit Swish hidden layer and 64 linear outputs, providing 32 scale and 32 shift coordinates. The FiLM strength is $\rho=0.10$. Unless an ablation explicitly changes the conditioning mechanism, the local optimizer architecture and structured update remain unchanged.

\subsection{Probe refresh and normalization}
\label{app:probe_refresh}
We used 16 PDE and 16 IC probe points to compute the dynamic PDE-state features. We sample these PDE probes using Latin hypercube sampling over the same spatio-temporal domain as the PDE residual points for each equation, and we sample the IC probes over the spatial domain at the initial time.  We compute residual concentration and roughness statistics from PDE residuals evaluated at the PDE probes, and update-response features from residuals evaluated at the PDE and IC probes. We compute the dynamic PDE-state features on the first optimizer call and refresh them every 50 calls. The first value initializes the exponential moving average; later refreshes are smoothed using
\begin{equation}
z_{\rm EMA}\leftarrow0.90z_{\rm EMA}+0.10z_{\rm new}.
\label{eq:pde_state_ema}
\end{equation}
We use seeded probe sampling in each experimental run for reproducibility. During meta-training, we derive probe seeds from the meta-training seed and task index. We use the test seed +777 at meta-test. In the feature study and the G2 conditioning validation, we use z-score normalization across the active feature coordinates smoothed by EMA at each refresh, while inactive coordinates are set to zero. All11 applies the same normalization across all eleven coordinates.

\subsection{PES meta-training}
\label{app:pes}

PES uses 20 tasks, one cycle, 50 outer steps, $K=1$, and $N=2$ particles forming one antithetic pair. Meta-training uses $\sigma=0.1$, outer learning rate $10^{-4}$, zero weight decay, and clipping. The implementation draws the positive perturbation using a Gaussian distribution with standard deviation $\sigma^2$; the second particle is its negative, and the accumulated gradient is divided by $N\sigma^2$. We use the same learned-optimizer meta-training configuration for the unconditioned baseline \BaseLO{} and the conditioned variants.

\subsection{Evaluation protocol}
\label{sec:training_evaluation}
We evaluate \Method{} against Adam and the matched unconditioned Bihlo-style meta-learned optimizer, denoted \BaseLO{}, using the same PINN architecture, training data, and evaluation protocol. For the learned optimizers, the principal comparison uses meta-training seeds $\{1234, 2024, 7 \}$. We evaluate each learned checkpoint for 1,000 epochs on all three PDEs with five meta-test runs. We first average these five meta-test runs for each learned checkpoint, then summarize the three meta-training means. We report relative $L_2$, L1, solution MSE, and physics MSE, where physics MSE is the sum of PDE-residual and IC-residual MSE on the evaluation grid.

\subsection{Design study protocol}
\label{sec:design_study}
We selected the final configuration (\Method{}) after targeted design studies on both state construction and the conditioning strategy. We separately varied \emph{what} PDE information we expose and \emph{how} we use it on the learned representation. The information study covers G1/G2/G3, all pairwise groups, and the full All11 state together with targeted residual-focused variants. For the conditioning study, we compared direct input (NoContext), gamma-only/beta-only FiLM, Groupwise Deconfounded FiLM, common-contrast variants, frozen branch masks, centering controls, and wider RawScale controllers. We then re-meta-trained promising alternatives independently to test whether their gains persisted across the stochastic optimizer learning process. The complete design study and results appear in Appendices~\ref{app:feature_groups} and \ref{app:conditioning}.


\section{PDE-state study}
\label{app:feature_groups}

\subsection{Feature-group screening}
\label{app:feature_screen}

We first meta-trained G1, G2, G3, all three pairwise combinations, and All11 once with meta-training seed 1234. Then, based on the relative $L_2$ performance, we selected top candidate groups for re-meta-training. Table~\ref{tab:group_relL2_full} reports the complete relative-$L_2$ screening history. Table~\ref{tab:group_allmetrics_agg} then summarizes all four endpoint metrics for the four configurations repeated at all three meta-training seeds. Keeping these two views separate distinguishes broad feature screening from the stronger independent-meta-training evidence.

\begin{table}[H]
\centering\small
\caption{PlainFiLM feature-group experiments. Values are relative-$L_2$ mean $\pm$ sample SD over five meta test runs for each fixed meta-trained optimizer. The complete seven-configuration screen is shown for meta-train-seed 1234; configurations repeated at meta-train-seeds 2024 and 7 are listed below it. Boldface marks the lowest mean within each meta-seed/PDE block.}
\label{tab:group_relL2_full}
\begin{tabular}{@{}r l ccc@{}}
\toprule
Meta seed & Active state & LA & KdV & Burgers\\
\midrule
1234 & G1 & .0962$\pm$.0426 & .3176$\pm$.0496 & .1601$\pm$.1039\\
& G2 & .1188$\pm$.0594 & $\mathbf{.2859\pm.0154}$ & $\mathbf{.0959\pm.0478}$\\
& G3 & .1143$\pm$.0641 & .3013$\pm$.0218 & .1529$\pm$.0575\\
& G1+G2 & .1216$\pm$.0610 & .3080$\pm$.0290 & .1151$\pm$.0454\\
& G1+G3 & .1185$\pm$.0453 & .3072$\pm$.0298 & .1035$\pm$.0431\\
& G2+G3 & $\mathbf{.0823\pm.0470}$ & .3309$\pm$.0563 & .1065$\pm$.0809\\
& All11 & .1698$\pm$.1517 & .3526$\pm$.0537 & .1061$\pm$.0543\\
\midrule
2024 & G2 & $\mathbf{.0569\pm.0280}$ & .3177$\pm$.0202 & .1499$\pm$.0677\\
& G1+G3 & .1171$\pm$.0991 & $\mathbf{.3081\pm.0800}$ & .1523$\pm$.0930\\
& G2+G3 & .1513$\pm$.0913 & .3189$\pm$.0599 & $\mathbf{.1209\pm.0824}$\\
& All11 & .1325$\pm$.0736 & .3170$\pm$.0474 & .1557$\pm$.0802\\
\midrule
7 & G2 & $\mathbf{.2259\pm.1226}$ & $\mathbf{.3727\pm.0679}$ & .1119$\pm$.0446\\
& G1+G3 & .2296$\pm$.1192 & .3803$\pm$.0683 & .0977$\pm$.0362\\
& G2+G3 & .2328$\pm$.0982 & .3899$\pm$.0665 & .1328$\pm$.0443\\
& All11 & .3127$\pm$.0580 & .4202$\pm$.0595 & $\mathbf{.0878\pm.0319}$\\
\bottomrule
\end{tabular}
\end{table}

\begin{table}[H]
\centering\scriptsize
\caption{Averaged four-metric results for the PlainFiLM state configurations repeated at meta-training seeds 1234, 2024, and 7. Five meta test runs are first averaged within each checkpoint; entries are mean $\pm$ sample SD across the three checkpoint means. Boldface marks the lowest mean among these repeated configurations for each PDE/metric.}
\label{tab:group_allmetrics_agg}
\begin{tabular}{@{}llcccc@{}}
\toprule
State & PDE & Rel. $L_2$ & L1 & Solution MSE & Physics MSE\\
\midrule
G2 & LA & $\mathbf{.1339\pm.0855}$ & $\mathbf{.0770\pm.0498}$ & $\mathbf{.0140\pm.0155}$ & $\mathbf{.0091\pm.0059}$\\
& KdV & $\mathbf{.3254\pm.0439}$ & $\mathbf{.1633\pm.0172}$ & $\mathbf{.0546\pm.0156}$ & .0211$\pm$.0017\\
& Burgers & .1192$\pm$.0277 & $\mathbf{.0304\pm.0027}$ & $\mathbf{.0064\pm.0030}$ & $\mathbf{.0065\pm.0013}$\\
\midrule
G1+G3 & LA & .1551$\pm$.0646 & .0893$\pm$.0377 & .0169$\pm$.0132 & .0105$\pm$.0042\\
& KdV & .3319$\pm$.0419 & .1647$\pm$.0164 & .0575$\pm$.0148 & $\mathbf{.0205\pm.0023}$\\
& Burgers & .1178$\pm$.0300 & .0331$\pm$.0049 & .0066$\pm$.0041 & .0095$\pm$.0037\\
\midrule
G2+G3 & LA & .1555$\pm$.0753 & .0889$\pm$.0444 & .0167$\pm$.0134 & .0110$\pm$.0048\\
& KdV & .3466$\pm$.0380 & .1706$\pm$.0140 & .0624$\pm$.0138 & .0211$\pm$.0021\\
& Burgers & .1201$\pm$.0132 & .0356$\pm$.0060 & .0070$\pm$.0007 & .0126$\pm$.0074\\
\midrule
All11 & LA & .2050$\pm$.0951 & .1175$\pm$.0558 & .0282$\pm$.0201 & .0141$\pm$.0065\\
& KdV & .3633$\pm$.0524 & .1787$\pm$.0218 & .0684$\pm$.0198 & .0228$\pm$.0045\\
& Burgers & $\mathbf{.1165\pm.0351}$ & .0365$\pm$.0053 & .0065$\pm$.0041 & .0092$\pm$.0028\\
\bottomrule
\end{tabular}
\end{table}

The screening and repeated runs show that the G2 feature group wins most equations or most meta-training runs.

\subsection{Residual-focused follow-up experiments}
\label{app:residual}
The recurring strength of G2 motivated targeted follow-ups that test whether adding a specific PDE--IC balance signal or imposing a more specialized residual transformation improves on the original residual-geometry state.

\paragraph{G2-ext.}
G2-ext expands the original two-dimensional G2 state with five additional
residual-focused coordinates. The added features are \emph{log PDE-loss norm}, \emph{PDE decay}, \emph{tail ratio}, \emph{log PDE--IC loss ratio}, and \emph{IC drift}. The purpose of this variant
is to test whether a richer description of residual magnitude, evolution, tail behavior, and PDE--IC imbalance provides useful information beyond $C_{\rm PDE}$ and $R_{\rm PDE}$.

\paragraph{G2-balance.}
G2-balance keeps $C_{\rm PDE}$ and $R_{\rm PDE}$ and adds a PDE--IC gradient-balance feature
\begin{equation}
b_{\rm grad}=\tanh\left(
\frac{1}{3}\,
\mathrm{clip}_{[-6,6]}
\left[
\log_{10}
\frac{\|\nabla_\theta\LPDE\|_2+\eps}
     {\|\nabla_\theta\LIC\|_2+\eps}
\right]\right).
\label{eq:gradient_balance}
\end{equation}
The gradients are calculated with respect to all trainable PINN parameters. We use clipping to limit extreme gradient-norm ratios before the bounded transformation. Positive $b_{\rm grad}$ indicates that the PDE-gradient norm is larger, negative values indicate that the IC-gradient norm is larger, and values near zero indicate comparable magnitudes. Then we apply EMA smoothing to the three feature values. After smoothing, we normalize $C_{\rm PDE}$ and $R_{\rm PDE}$ the same way as the original G2 feature normalization. The balance coordinate is clipped to $[-1,1]$ before being supplied to the context pathway.

\paragraph{ResGeomBasis.}
ResGeomBasis instead tests whether the interaction between the two G2 coordinates is more useful than treating them only as separate state variables.
Let $\widetilde C_{\rm PDE}$ and $\widetilde R_{\rm PDE}$ denote the normalized G2 coordinates. The variant constructs
\begin{equation}
q_{\rm RG}=
\begin{bmatrix}
\widetilde C_{\rm PDE} &
\widetilde R_{\rm PDE} &
\widetilde C_{\rm PDE}\widetilde R_{\rm PDE}
\end{bmatrix}^{\!\top},
\label{eq:resgeom_basis}
\end{equation}
and maps this three-dimensional basis to the 32-dimensional first hidden layer using a bias-free linear projection. Rather than the usual PlainFiLM scale-shift transformation, we apply the resulting signal as a bounded shift.
\begin{equation}
\widetilde h
=
h+0.10\,\tanh\!\left(W_{\rm RG}q_{\rm RG}\right).
\label{eq:resgeom_shift}
\end{equation}
This isolates whether explicitly representing the concentration--roughness interaction provides a more effective residual-geometry conditioning signal.

\begin{table}[H]
\centering\scriptsize
\caption{Residual-focused follow-ups. G2 is repeated within each corresponding block so that every comparison uses the correct checkpoint family. Values are the mean $\pm$ sample SD over five PINN test seeds; boldface marks the lower mean within each directly comparable block.}
\label{tab:residual_followups}
\begin{tabular}{@{}rllcccc@{}}
\toprule
Meta seed & Variant & PDE & Rel. $L_2$ & L1 & Solution MSE & Physics MSE\\
\midrule
1234 & G2 & LA & $\mathbf{.1188\pm.0594}$ & $\mathbf{.0677\pm.0333}$ & $\mathbf{.0085\pm.0076}$ & $\mathbf{.0085\pm.0037}$\\
& & KdV & $\mathbf{.2859\pm.0154}$ & $\mathbf{.1469\pm.0052}$ & $\mathbf{.0412\pm.0045}$ & $\mathbf{.0202\pm.0018}$\\
& & Burgers & $\mathbf{.0959\pm.0478}$ & $\mathbf{.0281\pm.0054}$ & $\mathbf{.0041\pm.0041}$ & $\mathbf{.0080\pm.0043}$\\
1234 & G2-ext & LA & .2422$\pm$.1471 & .1399$\pm$.0847 & .0380$\pm$.0355 & .0177$\pm$.0119\\
& & KdV & .3375$\pm$.0423 & .1685$\pm$.0191 & .0579$\pm$.0135 & .0217$\pm$.0031\\
& & Burgers & .1317$\pm$.0954 & .0325$\pm$.0143 & .0093$\pm$.0137 & .0108$\pm$.0090\\
\midrule
2024 & G2 & LA & $\mathbf{.0569\pm.0280}$ & $\mathbf{.0325\pm.0163}$ & $\mathbf{.0019\pm.0017}$ & $\mathbf{.0035\pm.0023}$\\
& & KdV & .3177$\pm$.0202 & .1619$\pm$.0063 & .0509$\pm$.0064 & .0200$\pm$.0016\\
& & Burgers & $\mathbf{.1499\pm.0677}$ & $\mathbf{.0334\pm.0094}$ & $\mathbf{.0098\pm.0071}$ & .0057$\pm$.0021\\
2024 & G2-balance & LA & .1310$\pm$.0621 & .0742$\pm$.0367 & .0101$\pm$.0070 & .0091$\pm$.0057\\
& & KdV & $\mathbf{.3127\pm.0592}$ & $\mathbf{.1582\pm.0261}$ & $\mathbf{.0506\pm.0192}$ & $\mathbf{.0185\pm.0043}$\\
& & Burgers & .1677$\pm$.0728 & .0337$\pm$.0067 & .0122$\pm$.0088 & $\mathbf{.0056\pm.0019}$\\
2024 & ResGeomBasis-G2 & LA & 32.8736$\pm$26.2809 & 23.2160$\pm$18.6029 & 816.5711$\pm$989.4744 & 913.7524$\pm$919.9102\\
& & KdV & 5.8822$\pm$2.7532 & 4.0803$\pm$1.9981 & 20.4337$\pm$14.6996 & $2.517\!\times\!10^{11}\pm5.620\!\times\!10^{11}$\\
& & Burgers & 2.7159$\pm$1.2403 & 1.5275$\pm$.7708 & 3.2379$\pm$3.0723 & $4.636\!\times\!10^{3}\pm6.351\!\times\!10^{3}$\\
\bottomrule
\end{tabular}
\end{table}

At seed 2024, G2-balance gives small improvements over G2 on all four KdV metrics, but the same modification is worse on the LA solution and physics metrics and does not improve Burgers solution error. Only Burgers physics MSE is marginally lower. ResGeomBasis performs poorly across every stored metric. These controls therefore suggest that G2's strength cannot be explained simply by adding more residual structure.

\subsection{Final G2 comparison}
\label{app:matched}
For completeness, Table~\ref{tab:perblock_allmetrics} reports the mean over five meta test runs for each matched evaluation block in the final G2 comparison. For the learned optimizers, each block corresponds to an independently
meta-trained checkpoint. We compute the aggregate Table~\ref{tab:headline_allmetrics} from these three block means. 

The table also makes the scope of the robustness claim explicit. The conditioned optimizer is not the lowest in every individual block. The main claim is a lower aggregate mean in all 12 PDE-metric comparisons and substantially lower variability between relearning runs.

\begin{table}[H]
\centering\scriptsize
\caption{Per-meta-training-seed means for Adam, the unconditioned learned optimizer, and G2+PlainFiLM. Every cell averages five PINN test seeds. Boldface marks the lowest mean within each meta-training-seed/PDE/metric comparison.}
\label{tab:perblock_allmetrics}
\begin{tabular}{@{}rllcccc@{}}
\toprule
Meta seed & Method & PDE & Rel. $L_2$ & $L_1$ & Solution MSE & Physics MSE\\
\midrule
1234 & Adam & LA & .5451&.3155&.14959&.04365\\
& & KdV & .4485&.2147&.10204&.03357\\
& & Burgers & .1267&.0410&.00630&.01025\\
& \BaseLO & LA & .5969&.3480&.18567&.05058\\
& & KdV & .5677&.2642&.16417&.04930\\
& & Burgers & .3045&.1150&.03811&.05450\\
& \Method{} & LA & $\mathbf{.1188}$&$\mathbf{.0677}$&$\mathbf{.00847}$&$\mathbf{.00852}$\\
& & KdV & $\mathbf{.2859}$&$\mathbf{.1469}$&$\mathbf{.04118}$&$\mathbf{.02021}$\\
& & Burgers & $\mathbf{.0959}$&$\mathbf{.0281}$&$\mathbf{.00414}$&$\mathbf{.00804}$\\
\midrule
2024 & Adam & LA & .5123&.2954&.13284&.04025\\
& & KdV & .4769&.2258&.11549&.03695\\
& & Burgers & .1292&.0447&.00675&.01220\\
& \BaseLO & LA & .2365&.1369&.04004&.01727\\
& & KdV & .3455&.1727&.06290&.02514\\
& & Burgers & $\mathbf{.0751}$&$\mathbf{.0293}$&$\mathbf{.00219}$&.00956\\
& \Method{} & LA & $\mathbf{.0569}$&$\mathbf{.0325}$&$\mathbf{.00193}$&$\mathbf{.00354}$\\
& & KdV & $\mathbf{.3177}$&$\mathbf{.1619}$&$\mathbf{.05088}$&$\mathbf{.02005}$\\
& & Burgers & .1499&.0334&.00983&$\mathbf{.00565}$\\
\midrule
7 & Adam & LA & .5025&.2913&.14019&.04079\\
& & KdV & .4690&.2228&.11101&.03516\\
& & Burgers & .1300&.0419&.00665&.01062\\
& \BaseLO & LA & .4412&.2555&.12026&.03569\\
& & KdV & .4045&.1960&.08303&.02691\\
& & Burgers & $\mathbf{.0913}$&.0302&$\mathbf{.00323}$&$\mathbf{.00528}$\\
& \Method{} & LA & $\mathbf{.2259}$&$\mathbf{.1307}$&$\mathbf{.03153}$&$\mathbf{.01517}$\\
& & KdV & $\mathbf{.3727}$&$\mathbf{.1812}$&$\mathbf{.07165}$&$\mathbf{.02296}$\\
& & Burgers & .1119&$\mathbf{.0296}$&.00531&.00594\\
\bottomrule
\end{tabular}
\end{table}

\section{Conditioning-mechanism study}
\label{app:conditioning}
We developed the conditioning study in two stages. We first used the full All11 PDE state as a fixed information interface and varied how that state entered the learned optimizer. This broader study included direct input, scale/shift ablations, common-contrast variants, frozen-branch diagnostics, capacity-matched centering controls, Groupwise Deconfounded FiLM, and wider RawScale controllers. Most of the mechanism controls reported in this appendix therefore use All11. These experiments aimed to isolate whether the gains came from state exposure itself, a particular affine component, hidden-state centering, semantic grouping, or controller capacity. The
corresponding fixed-checkpoint and mechanism-control results are reported in the following subsections, including the structured-modulation comparison in Table~\ref{tab:fixed_checkpoint_modulation}.

The later PDE-state study identified G2 as the strongest overall state representation. We therefore performed a targeted conditioning mechanism cross-check, keeping G2 fixed rather than repeating the entire conditioning sweep. PlainFiLM was retained as the reference conditioning mechanism, ContrastOnly was carried forward because of its strong fixed-checkpoint behavior with All11, and RawScale96 served as the higher-capacity alternative. All three were independently re-meta-trained with meta-seeds 1234, 2024, and 7. 

Table~\ref{tab:conditioning_g2_full} reports this final G2-fixed comparison, which is the basis for the conditioning conclusion in the main paper. For context, in Table~\ref{tab:conditioning_all11} we report the corresponding earlier repeated study with All11 fixed. 

Under All11, PlainFiLM is lower on LA and Burgers, whereas RawScale96 gives the lowest aggregate mean on KdV and ContrastOnly does not retain its earlier fixed-checkpoint advantage. With G2 fixed, PlainFiLM is lower than both ContrastOnly and RawScale96 on all PDE--metric comparisons. This comparison shows that the preferred conditioning mechanism is not completely independent of the PDE-state representation, motivating the targeted G2-fixed validation rather than assuming that the All11 ranking would carry over unchanged.

\begin{table}[H]
\centering\scriptsize
\caption{Independent conditioning comparison with G2 fixed. Five meta-test runs are averaged within each checkpoint; entries are the mean $\pm$ sample SD across three meta-training runs. Boldface marks the lowest mean for each PDE/metric.}
\label{tab:conditioning_g2_full}
\begin{tabular}{@{}llcccc@{}}
\toprule
Conditioning & PDE & Rel. $L_2$ & $L_1$ & Solution MSE & Physics MSE\\
\midrule
PlainFiLM & LA & $\mathbf{.1339\pm.0855}$ & $\mathbf{.0770\pm.0497}$ & $\mathbf{.01398\pm.01555}$ & $\mathbf{.00908\pm.00584}$\\
& KdV & $\mathbf{.3255\pm.0439}$ & $\mathbf{.1633\pm.0172}$ & $\mathbf{.05457\pm.01557}$ & $\mathbf{.02107\pm.00164}$\\
& Burgers & $\mathbf{.1192\pm.0278}$ & $\mathbf{.0304\pm.0027}$ & $\mathbf{.00643\pm.00301}$ & $\mathbf{.00655\pm.00130}$\\
\midrule
ContrastOnly & LA & .2890$\pm$.1659 & .1663$\pm$.0966 & .05596$\pm$.05199 & .02192$\pm$.01305\\
& KdV & .4083$\pm$.0561 & .1976$\pm$.0228 & .08889$\pm$.02495 & .03164$\pm$.01035\\
& Burgers & .2079$\pm$.0930 & .0650$\pm$.0438 & .02350$\pm$.02372 & .02652$\pm$.02899\\
\midrule
RawScale96 & LA & .2962$\pm$.1548 & .1707$\pm$.0899 & .05620$\pm$.04878 & .02269$\pm$.01205\\
& KdV & .3829$\pm$.0451 & .1856$\pm$.0172 & .07546$\pm$.01898 & .02439$\pm$.00448\\
& Burgers & .1256$\pm$.0087 & .0355$\pm$.0017 & .00679$\pm$.00061 & .00824$\pm$.00162\\
\bottomrule
\end{tabular}
\end{table}

\begin{table}[H]
\centering\scriptsize
\caption{Earlier independent conditioning comparison with the All11 PDE state fixed. Values are the mean $\pm$ sample SD across three checkpoint means after averaging five meta-test seeds within each checkpoint; boldface marks the lowest mean for each PDE/metric.}
\label{tab:conditioning_all11}
\begin{tabular}{@{}llcccc@{}}
\toprule
Conditioning & PDE & Rel. $L_2$ & L1 & Solution MSE & Physics MSE\\
\midrule
PlainFiLM & LA & $\mathbf{.1365\pm.0896}$ & $\mathbf{.0781\pm.0522}$ & $\mathbf{.0153\pm.0139}$ & $\mathbf{.0092\pm.0053}$\\
& KdV & $.3445\pm.0320$ & $.1717\pm.0129$ & $.0612\pm.0110$ & $.0222\pm.0028$\\
& Burgers & $\mathbf{.1407\pm.0395}$ & $\mathbf{.0344\pm.0056}$ & $\mathbf{.0091\pm.0053}$ & $\mathbf{.0085\pm.0009}$\\
ContrastOnly & LA & $.2933\pm.1569$ & $.1688\pm.0917$ & $.0560\pm.0493$ & $.0222\pm.0128$\\
& KdV & $.3957\pm.0565$ & $.1909\pm.0219$ & $.0817\pm.0254$ & $.0263\pm.0064$\\
& Burgers & $.2120\pm.1485$ & $.0718\pm.0601$ & $.0311\pm.0409$ & $.0348\pm.0430$\\
RawScale96 & LA & $.2281\pm.2294$ & $.1308\pm.1330$ & $.0461\pm.0613$ & $.0177\pm.0187$\\
& KdV & $\mathbf{.3289\pm.1177}$ & $\mathbf{.1616\pm.0513}$ & $\mathbf{.0602\pm.0386}$ & $\mathbf{.0210\pm.0100}$\\
& Burgers & $.1746\pm.1134$ & $.0529\pm.0418$ & $.0195\pm.0245$ & $.0211\pm.0267$\\
\bottomrule
\end{tabular}
\end{table}

\subsection{Direct-input control}
\label{app:nocontext}
The NoContext control experiment answers whether simply exposing PDE-level information to the parameter-wise optimizer is sufficient without separate conditioning. In the direct-input NoContext control, selected PDE-state coordinates are appended to the meta-learned optimizer inputs before the first hidden layer. Table~\ref{tab:nocontext_results} compares the NoContext experiment runs with the corresponding PlainFiLM feature-group results:
\begin{table}[H]
\centering\small
\caption{Relative $L_2$ performance of Direct-input NoContext versus PlainFiLM with the same active PDE-state features. All optimizers use meta-training seed 1234; values are mean $\pm$
sample SD over the same five meta-test seeds. The checkpoints are independently meta-trained. Lower is better; boldface marks the lower mean within each state/PDE comparison.}
\label{tab:nocontext_results}
\begin{tabular}{@{}llccc@{}}
\toprule
State & Method & LA & KdV & Burgers\\
\midrule
G1 & PlainFiLM & $\mathbf{.0962\pm.0426}$ & $\mathbf{.3176\pm.0496}$ & $\mathbf{.1601\pm.1039}$\\
& NoContext & .1622$\pm$.0572 & .5864$\pm$.2750 & .3224$\pm$.1020\\
\midrule
G2 & PlainFiLM & .1188$\pm$.0594 & $\mathbf{.2859\pm.0154}$ & $\mathbf{.0959\pm.0478}$\\ 
& NoContext & $\mathbf{.0793\pm.0188}$ & .6041$\pm$.1874 & .6442$\pm$.2613\\
\midrule
G3 & PlainFiLM & $\mathbf{.1143\pm.0641}$ & $\mathbf{.3013\pm.0218}$ & $\mathbf{.1529\pm.0575}$\\
& NoContext & .3479$\pm$.2817 & .7903$\pm$.5344 & .6241$\pm$.2992\\
\midrule
All11 & PlainFiLM & $\mathbf{.1698\pm.1517}$ & $\mathbf{.3526\pm.0537}$ & $\mathbf{.1061\pm.0543}$\\
& NoContext & .8944$\pm$.0725 & .7435$\pm$.3858 & .5443$\pm$.1795\\
\bottomrule
\end{tabular}
\end{table}
These runs show that providing the global PDE state as a direct input does not guarantee reliable transfer on its own. NoContext-G2 is better on LA than PlainFiLM. But this performance reverses sharply on KdV and Burgers. The remaining NoContext states are worse than their corresponding PlainFiLM variants across all three PDEs. Because the PlainFiLM and NoContext checkpoints are independently meta-trained, we treat this as supporting evidence that information placement matters rather than as a causal proof that FiLM is necessary.

\subsection{Scale and shift ablations}
\label{app:scale_shift}
The performance gain of PlainFiLM conditioning led us to further explore whether the gain comes from the multiplicative scale or from the additive shift. GammaOnlyFiLM keeps the bounded multiplicative scale and removes the shift. Its fixed-checkpoint five-test-seed relative-$L_2$ means are $0.1691\pm0.1689$, $0.3414\pm0.0659$, and $0.0987\pm0.0521$ on LA/KdV/ Burgers, compared to $0.0969\pm0.0614$, $0.3212\pm0.0431$, and $0.0871\pm0.0149$ for the corresponding PlainFiLM checkpoint. BetaOnlyFiLM was evaluated on one complete endpoint test run. On that run, it improves LA relative-$L_2$ over PlainFiLM but is worse on KdV and Burgers. The evidence therefore does not support either scale-only or shift-only as a general replacement for the full bounded affine conditioning.

\subsection{Common--contrast decomposition}
\label{app:common_contrast}
PlainFiLM applies the PDE-conditioned affine transformation directly to each parameter hidden representation similarly. We therefore asked whether it would help to separate information shared across all PINN parameters from parameter-specific deviations around that shared state. The motivation was that a global PDE signal may not need to modify every part of the hidden representation in the same way. It could act on a common optimizer tendency, on how individual parameters deviate from that tendency, or on both. To test this, we decompose the first hidden representation across scalar PINN parameters into a common and contrast part like the following:
\begin{equation}
\mu_H=\frac1P\sum_{i=1}^P h_i,\qquad \delta_i=h_i-\mu_H,
\label{eq:common_contrast}
\end{equation}
where $\mu_H$ is the common component and $\delta_i$ is the parameter-specific contrast. Hence, the common--contrast experiment acts on the common component, contrast, shift, or selected combinations of these branches. 
Table~\ref{tab:fixed_checkpoint_modulation} reports the fixed-checkpoint comparison of PlainFiLM, GammaOnlyFiLM, and the common--contrast variants with All11 fixed. ContrastOnly is the strongest variant on several LA and Burgers metrics, which motivated its further study. In contrast, Full CommonContrast and CommonOnly are worse than PlainFiLM across the three solution-error metrics.
However, the repeated All11 comparison in Table~\ref{tab:conditioning_all11} shows that the apparent ContrastOnly advantage does not persist after independently relearning the optimizer. The G2-fixed validation in Table~\ref{tab:conditioning_g2_full} reaches the same conclusion with the selected state held fixed.


\begin{table}[H]
\centering\scriptsize
\caption{Complete fixed-checkpoint structured-modulation endpoint metrics with All11 fixed. Values are mean $\pm$ sample SD over five test seeds for one meta-trained checkpoint per method; boldface marks the lowest mean for each PDE/metric (ties are both bold).}
\label{tab:fixed_checkpoint_modulation}
\begin{tabular}{@{}llcccc@{}}
\toprule
Method & PDE & Rel. $L_2$ & L1 & Solution MSE & Physics MSE\\
\midrule
PlainFiLM & LA & .0969$\pm$.0614 & .0542$\pm$.0363 & .0062$\pm$.0073 & .0078$\pm$.0048\\
& KdV & $\mathbf{.3212\pm.0431}$ & $\mathbf{.1589\pm.0159}$ & $\mathbf{.0526\pm.0143}$ & $\mathbf{.0188\pm.0020}$\\
& Burgers & .0871$\pm$.0149 & .0279$\pm$.0050 & .0029$\pm$.0009 & .0122$\pm$.0094\\
Gamma-only & LA & .1691$\pm$.1689 & .0971$\pm$.0972 & .0257$\pm$.0373 & .0114$\pm$.0130\\
& KdV & .3414$\pm$.0659 & .1692$\pm$.0264 & .0603$\pm$.0236 & .0217$\pm$.0059\\
& Burgers & .0987$\pm$.0521 & .0264$\pm$.0051 & .0045$\pm$.0049 & $\mathbf{.0032\pm.0014}$\\
Full common--contrast & LA & .2178$\pm$.1318 & .1245$\pm$.0769 & .0307$\pm$.0280 & .0158$\pm$.0105\\
& KdV & .3907$\pm$.0262 & .1905$\pm$.0095 & .0770$\pm$.0101 & .0260$\pm$.0034\\
& Burgers & .1561$\pm$.0624 & .0351$\pm$.0055 & .0103$\pm$.0078 & .0084$\pm$.0029\\
Common-only & LA & .1841$\pm$.1035 & .1052$\pm$.0602 & .0212$\pm$.0208 & .0125$\pm$.0067\\
& KdV & .3600$\pm$.0839 & .1768$\pm$.0333 & .0679$\pm$.0321 & .0249$\pm$.0069\\
& Burgers & .1046$\pm$.0328 & .0300$\pm$.0035 & .0044$\pm$.0027 & .0093$\pm$.0025\\
Contrast-only & LA & $\mathbf{.0820\pm.0448}$ & $\mathbf{.0471\pm.0264}$ & $\mathbf{.0042\pm.0040}$ & $\mathbf{.0043\pm.0027}$\\
& KdV & .3255$\pm$.0518 & .1642$\pm$.0201 & .0543$\pm$.0183 & $\mathbf{.0188\pm.0024}$\\
& Burgers & $\mathbf{.0779\pm.0312}$ & $\mathbf{.0229\pm.0034}$ & $\mathbf{.0026\pm.0022}$ & .0033$\pm$.0016\\
Contrast+shift & LA & .2184$\pm$.1283 & .1261$\pm$.0739 & .0304$\pm$.0304 & .0153$\pm$.0089\\
& KdV & .3434$\pm$.0359 & .1681$\pm$.0152 & .0598$\pm$.0123 & .0200$\pm$.0038\\
& Burgers & .1039$\pm$.0442 & .0307$\pm$.0056 & .0047$\pm$.0037 & .0081$\pm$.0012\\
\bottomrule
\end{tabular}
\end{table}

\subsection{Capacity-matched centering control}
\label{app:centering}
The strong fixed-checkpoint performance of ContrastOnly raised another, simpler possibility. Maybe its advantage came mainly from centering the hidden representation, rather than from the contrast-based conditioning itself. To check that possibility, we compare two capacity-matched controllers. RawScale32 and ContrastOnly32 use the same 32-output context network and the same modulation strength. They differ only in whether the bounded scale acts directly on the hidden representation $h_i$ or on the centered deviation $h_i-\mu_H$. Their relative-$L_2$ results are:

\begin{table}[H]
\centering\small
\caption{Capacity-matched centering control (meta-seed 1234; five meta-test runs). Boldface marks the lower mean for each PDE.}
\label{tab:centering_control}
\begin{tabular}{@{}lccc@{}}
\toprule
Method & LA & KdV & Burgers\\
\midrule
RawScale32 & $\mathbf{.0792\pm.0488}$ & $\mathbf{.3214\pm.0423}$ & .1171$\pm$.0621\\
ContrastOnly32 & .1188$\pm$.0807 & .3742$\pm$.0421 & $\mathbf{.1038\pm.0535}$\\
\bottomrule
\end{tabular}
\end{table}
RawScale32 is lower on LA/KdV, while ContrastOnly32 has a lower Burgers mean but wins only two of five paired Burgers relative-$L_2$ tests. Mean-centering alone therefore does not explain the earlier ContrastOnly checkpoint.

\subsection{Groupwise Deconfounded FiLM}
\label{app:groupwise}
Groupwise Deconfounded FiLM sends G1, G2, and G3 through separate context heads and adds their bounded scale/shift contributions. It provides a strong seed-1234 checkpoint, but does not reproduce the LA/KdV gains at seed 2024.
\begin{table}[H]
\centering\small
\caption{Groupwise Deconfounded FiLM versus PlainFiLM-All11. Values are relative-$L_2$ mean $\pm$ sample SD over five meta-test runs; boldface marks the lowest mean within each meta-seed/PDE block.}
\label{tab:groupwise_film}
\begin{tabular}{@{}r l ccc@{}}
\toprule
Meta seed & Method & LA & KdV & Burgers\\
\midrule
1234 & PlainFiLM-All11 & .1698$\pm$.1517 & .3526$\pm$.0537 & .1061$\pm$.0543\\
& Groupwise FiLM & $\mathbf{.1262\pm.0747}$ & $\mathbf{.2623\pm.0347}$ & $\mathbf{.0953\pm.0465}$\\
2024 & PlainFiLM-All11 & $\mathbf{.1325\pm.0736}$ & $\mathbf{.3170\pm.0474}$ & .1557$\pm$.0802\\
& Groupwise FiLM & .5705$\pm$.0575 & .4807$\pm$.0383 & .1055$\pm$.0098\\
& Groupwise FiLM (same norm) & .5623$\pm$.0732 & .4835$\pm$.0388 & $\mathbf{.1007\pm.0281}$\\
\bottomrule
\end{tabular}
\end{table}

A normalization-matched control remains poor on LA/KdV, so the seed-2024 reversal is not repaired by that normalization switch.

\subsection{Higher-capacity control: RawScale96}
\label{app:rawscale_checkpoints}
We also considered a separate higher-capacity alternative, RawScale96, to test whether a wider context controller with direct multiplicative modulation could provide more effective conditioning without contrast decomposition. RawScale96 uses a wider 96-output context controller to produce a parameter-wise multiplicative signal $c(s_t)$ from the PDE state. Rather than forming separate bounded scale and shift terms as in PlainFiLM, the controller
acts directly on the hidden representation,
\begin{equation}
\widetilde h_i
=
h_i+c(s_t)\odot h_i
=
\bigl(1+c(s_t)\bigr)\odot h_i .
\label{eq:rawscale96}
\end{equation}
Thus, RawScale96 tests whether increased controller capacity, together with direct multiplicative modulation, is sufficient to achieve strong conditioning without either the contrast decomposition or the additive shift used by PlainFiLM.
Table~\ref{tab:conditioning_per_checkpoint} expands the All11 aggregate comparison in Table~\ref{tab:conditioning_all11} by reporting the individual checkpoint means used to compute the aggregate mean and SD.
\begin{table}[H]
\centering\small
\caption{Per-checkpoint relative-$L_2$ for the conditioning-ablation experiment. Boldface marks the lowest mean within each meta-training-seed/PDE block.}
\label{tab:conditioning_per_checkpoint}
\begin{tabular}{@{}r l ccc@{}}
\toprule
Meta seed & Method & LA & KdV & Burgers\\
\midrule
1234 & PlainFiLM & .0411 & .3459 & .0971\\
& ContrastOnly & .2547 & .3604 & .0988\\
& RawScale96 & $\mathbf{.0277}$ & $\mathbf{.2052}$ & $\mathbf{.0827}$\\
2024 & PlainFiLM & $\mathbf{.1494}$ & $\mathbf{.3118}$ & $\mathbf{.1512}$\\
& ContrastOnly & .1592 & .3657 & .3801\\
& RawScale96 & .1784 & .3422 & .3014\\
7 & PlainFiLM & $\mathbf{.2189}$ & $\mathbf{.3757}$ & .1739\\
& ContrastOnly & .4659 & .4609 & .1570\\
& RawScale96 & .4782 & .4395 & $\mathbf{.1397}$\\
\bottomrule
\end{tabular}
\end{table}
This table clearly illustrates why independent meta-training is necessary. RawScale96 looks dominant at seed 1234 and does not retain that behavior at the other two seeds. Similarly, the strong ContrastOnly checkpoint observed in the earlier fixed-checkpoint study does not persist after multiple meta-learning optimizer runs.

\end{document}